\documentclass[runningheads]{llncs}

\usepackage{eccv}

\usepackage{eccvabbrv}

\usepackage{graphicx}
\usepackage{booktabs}
\usepackage{amsmath}
\usepackage{array}
\usepackage{float}
\usepackage{placeins}

\usepackage[accsupp]{axessibility}  

\usepackage{hyperref}

\usepackage{orcidlink}

\begin{document}

\title{Style or Signature? Artist-Disjoint Evaluation of Style Classification in Frozen Vision Embeddings}
\titlerunning{Artist-Disjoint Style Classification}
\author{Rory Ashton}
\authorrunning{Rory Ashton}
\institute{University of Edinburgh \\ \email{R.P.Ashton@sms.ed.ac.uk}}
\maketitle

\begin{abstract}
Frozen image embeddings from models such as CLIP are increasingly used to classify paintings by art-historical style, with high reported accuracy. We ask whether this accuracy reflects an understanding of style or the recognition of individual artists. Standard evaluation uses random splits in which works by the same artist appear on both sides, so a classifier can succeed by recognising the painter rather than the movement. We re-evaluate style classification under an artist-disjoint protocol, holding out every artist in turn so that no work is ever classified using other works by its own painter. On a balanced dataset of 320 paintings across four twentieth-century movements, 5-NN style accuracy falls from 0.87 to 0.77 under this protocol, and the drop is sharply uneven. Impressionism and Cubism barely move, while Surrealism falls twenty points. The pattern holds across four image encoders, including a vision-only self-supervised model, which places the effect in visual structure rather than language. Where an encoder captures genuine shared form, individual artists are barely recognisable yet style is robust, while Surrealism shows the opposite. We argue that artist-disjoint evaluation is necessary to measure stylistic understanding in frozen embeddings.
\end{abstract}


\section{Introduction}
\label{sec:intro}

Frozen image embeddings from large pretrained vision models, CLIP~\cite{radford2021clip} foremost among them, are increasingly used off the shelf to analyse visual art, including organising collections, retrieving similar works, and classifying paintings by art-historical style~\cite{asperti2025clip,ghildyal2025wpclip}. A common and encouraging finding is that style is highly recoverable from these embeddings, since a simple nearest-neighbour or linear classifier on frozen features can assign movement labels with high accuracy, as we ourselves find (Section~\ref{sec:fullpool}). It is tempting to read such accuracy as evidence that these encoders have learned something about artistic style as a by-product of web-scale pretraining.

We argue that this reading is premature, because the standard way of measuring it cannot distinguish style from a confound these embeddings encode strongly, the identity of the individual artist. Style-classification studies typically evaluate on a random split of works, in which paintings by a given artist may fall on both the reference and the query side~\cite{lecoutre2017recognizing}. A classifier can then label a new Dal\'i as Surrealist either because it recognises Surrealism or because it recognises Dal\'i, and the evaluation cannot tell which. This would be hypothetical if artist identity were weakly represented, but it is not. Frozen CLIP features support artist identification at high accuracy, a signal strong enough to have been described as a neural signature~\cite{moayeri2025rethinking}. Whenever same-artist works are available as neighbours, a style score may therefore be measuring artist recognition in part, and a single aggregate accuracy gives no way to see how much, or for which movements.

The remedy is to control for artist identity directly, evaluating on artist-disjoint splits in which no work is classified using other works by its own painter. This is standard hygiene in other domains, where models are tested on held-out patients or speakers to prevent identity leakage from being mistaken for task performance, and a recent survey of style classification recommends artist- or source-disjoint splits and examining \emph{why} a style label is assigned rather than only whether a model is accurate~\cite{styleclass_survey2026}. To our knowledge this control has not been applied to style classification in frozen vision embeddings. We apply it and find that it changes the conclusions substantially.

On a balanced dataset of 320 paintings spanning four twentieth-century movements, replacing the standard evaluation with an artist-disjoint one lowers 5-NN style accuracy from $0.87$ to $0.77$. The aggregate drop is modest but averages over a sharply uneven effect. Impressionism and Cubism barely move, while Surrealism falls twenty points, near the level at which the model misclassifies half its works (Section~\ref{sec:loao}). An artist-recognisability control explains the unevenness (Section~\ref{sec:mechanism}). The robust movements are those whose individual artists the embedding can least tell apart, the mark of a shared visual form that has subsumed individual authorship, whereas Surrealism combines higher artist-recognisability with the lowest robustness, its collapse concentrated in specific painters whose style scores were borrowing from their recognisability as individuals. The pattern is not specific to CLIP. It holds across four encoders differing in scale, architecture, and training objective, including a vision-only model that rules out a language-based explanation (Section~\ref{sec:backbones}).

\paragraph{Contributions.}
\begin{enumerate}
  \item We introduce an \textbf{artist-disjoint (leave-one-artist-out) evaluation protocol} for painting-style classification in frozen vision embeddings, holding whole artists out of the neighbour pool, and show that it exposes fragility which standard random-split evaluation conceals.
  \item We quantify the gap between standard and artist-disjoint accuracy, show that it is \textbf{sharply movement-dependent} rather than uniform, and identify its mechanism with an \textbf{artist-recognisability control} that separates genuine shared form from artist memorisation, and further separates memorised artists from those the embedding encodes poorly.
  \item We show the effect is \textbf{encoder-general} across four image encoders, including one trained without language, which rules out a text-side explanation.
\end{enumerate}


\section{Related work}
\label{sec:related}

\paragraph{Frozen vision embeddings for art.}
The premise behind off-the-shelf use of encoders such as CLIP~\cite{radford2021clip} on paintings is that web-scale pretraining yields features that transfer to art without adaptation. Recent work probes this premise. Asperti et al.~\cite{asperti2025clip} report that CLIP captures semantic content more readily than stylistic nuance, and work on predicting W\"olfflin's principles finds that off-the-shelf CLIP struggles with fine-grained style and requires fine-tuning~\cite{ghildyal2025wpclip}. We ask a different question, not how well frozen embeddings classify style in absolute terms, but whether that accuracy measures style at all once the confounding contribution of artist identity is removed.

\paragraph{Style classification on WikiArt.}
Classifying paintings by movement is an established task, typically posed on WikiArt and evaluated by training a classifier and testing on a held-out split~\cite{lecoutre2017recognizing}. A recent survey observes that many studies train and evaluate within the same dataset, making it hard to tell whether models learn transferable stylistic features or dataset-specific artefacts, and calls for duplicate removal, artist- or source-disjoint splits, and transparent class distributions~\cite{styleclass_survey2026}. Our protocol implements that recommendation. Closest to our question, Strafforello et al.~\cite{strafforello2025art} evaluate zero-shot classification of style, author and period by CLIP, LLaVA and GPT-4o on WikiArt at scale, probing the models through their language interface. Two of their observations anticipate ours. CLIP frequently misclassifies Surrealism as Analytical Cubism, matching one of the confusions we find, and Picasso is not among CLIP's most predicted authors despite his prevalence in WikiArt, consistent with our finding that Cubist artists are the hardest for the embedding to tell apart.

\paragraph{Artist identity as a confound.}
That artist identity is strongly encoded in these representations is well established. Moayeri et al.~\cite{moayeri2025rethinking} show that a classifier on frozen features matches works to their artist for $89.3\%$ of $372$ WikiArt artists, describing these recurring characteristics as  \emph{neural signatures}. This is the signal a random-split style evaluation leaves uncontrolled.

\paragraph{Group-disjoint evaluation.}
Holding out a grouping variable to prevent identity leakage is established practice in adjacent fields. In medical imaging, the standard practice is to split at the patient level, so that a model cannot appear to detect a condition when it is actually recognising a patient it has already seen~\cite{rouzrokh2022mitigating}. In speech recognition, the common strategy is to hold out whole speakers, with the observation that which speaker is held out can drive substantial variability~\cite{liu2023partitioning}, a parallel to the per-artist variation we report. We apply this principle to painting-style classification in frozen vision embeddings.


\section{Data}
\label{sec:data}

We use a \emph{style dataset} of 320 paintings spanning four twentieth-century movements, built to study how cleanly frozen image embeddings separate art-historical styles and whether that separation survives when whole artists are held out.

\subsection{Style dataset}

\begin{figure}[t]
  \centering
  \includegraphics[width=0.24\linewidth]{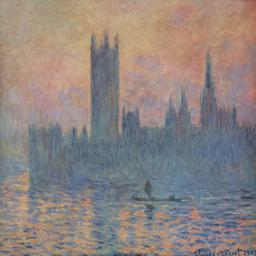}\hfill
  \includegraphics[width=0.24\linewidth]{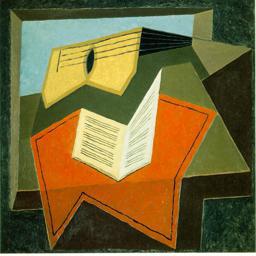}\hfill
  \includegraphics[width=0.24\linewidth]{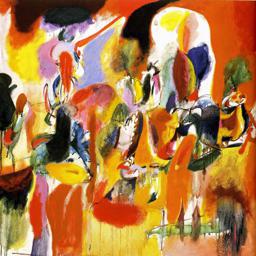}\hfill
  \includegraphics[width=0.24\linewidth]{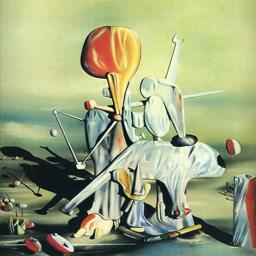}
  \caption{One work per movement from our dataset, shown as the $256\times256$ anisotropically resized thumbnails the encoders receive. Left to right, a Monet view built from visible strokes of colour (Impressionism), a Gris still life divided into flat geometric planes (Cubism), a Gorky composition of loose abstract forms (Abstract Expressionism), and a Tanguy scene of invented objects in deep empty space (Surrealism).}\label{fig:examples}
\end{figure}

\paragraph{Source and shape.}
The style dataset is drawn from the \texttt{Artificio\slash WikiArt} release on Hugging Face, a static snapshot of WikiArt. We sample from a frozen, versioned release rather than scraping the live site, so that the precise set is reproducible and cannot drift between experiments, and because this release assigns a single style label per work, avoiding the multi-tagging inflation of the live site where one work (Picasso's especially) may carry several movement tags. The subset is balanced by construction, with four movements, eight artists per movement, and ten works per artist, giving 320 images. Figure~\ref{fig:examples} shows one work from each movement. Our central experiment holds out whole artists (Section~\ref{sec:method}), so the more \emph{artists} per class, the more reliable each movement's accuracy. Giving every artist the same number of works also stops any one of them dominating a movement's result.

\paragraph{Artist selection.}
Each movement is represented by the eight artists most strongly associated with it in standard art-historical references~\cite{tate_glossary}, filtered to those with at least ten qualifying works for that movement in this release. We cite the selection \emph{criterion} rather than each artist, since the uncontested core of each movement (such as Monet, Picasso, and Dal\'i) follows from any standard reference. We tested an automatic ranking of candidates by tagged-work count and rejected it, since that count tracks digitisation volume rather than centrality, placing Sam Francis above Pollock and admitting Matisse to Impressionism. Work count is therefore used only as an inclusion filter. Four decisions are documented separately under \emph{Margin decisions}, either because canon is contested or because this release constrained the choice.

The eight artists per movement are Pollock, Rothko, Krasner, Kline, Newman, Still, Hofmann and Gorky (Abstract Expressionism); Picasso, Gris, L\'eger, Braque, Metzinger, Gleizes, Marcoussis and Villon (Cubism); Monet, Renoir, Degas, Sisley, Pissarro, Cassatt, Morisot and Caillebotte (Impressionism); and Dal\'i, Magritte, Ernst, Mir\'o, Delvaux, Masson, Carrington and Tanguy (Surrealism).

\paragraph{Work selection.}
For each artist we drew ten works at random from those tagged with any label in the movement's label set, above a resolution floor and with a fixed per-artist seed. In practice this release stores uniformly sized thumbnails, so the floor had no effect and every eligible work had the same chance of being drawn.

\paragraph{Label sets.}
Two of the four movements are umbrella terms that this release splits into sub-styles, so in each case we merge the sub-styles into a single label. Cubism merges Cubism, Analytical Cubism and Synthetic Cubism, which are phases of one movement. Abstract Expressionism merges Abstract Expressionism, Action Painting and Color Field Painting, the movement's two recognised wings alongside the general label~\cite{tate_abstractexpressionism}. Because we select artists first and then draw their works, merging Color Field into Abstract Expressionism admits first-generation figures such as Newman and Still without admitting the second-generation post-painterly painters (Louis, Noland, Olitski) who carry the same tag. Gating by artist does mean a movement is partly defined by our artist selection.

\paragraph{Preprocessing.}
Images are stored as $256\times256$ thumbnails produced by anisotropic resizing, a non-uniform squeeze that distorts aspect ratio but crops no content, so framing, composition and the figure-to-ground boundaries that matter for formal style are retained. Every painting is squeezed in the same way, and every encoder in our comparison receives the same input. The distortion does mean the images are not quite what these encoders saw at pretraining time.

\paragraph{Near-duplicate check.}
WikiArt occasionally stores a work twice, or a detail crop alongside a full reproduction, so we screened for duplicates, embedding all 320 images with CLIP ViT-B/32 and computing cosine similarity for every pair. No pair reached the $0.95$ flagging threshold. The closest pair (two works by Joan Mir\'o, cosine $0.949$) and the next-closest band (works by Gorky, Rothko and Pollock at $0.93$ to $0.94$) were inspected and confirmed distinct rather than duplicates.

\paragraph{Data availability.}
The 320 works are pinned by the \texttt{filename} field of the
\texttt{Artificio\slash WikiArt} release, recorded per image alongside its
SHA-256 content hash in the manifest and checksum list released with our
code and data at \url{https://github.com/Rory-A/ml-art-representation},
so the exact set is reproducible regardless of the selection procedure. Reconstruction must use this release, since an attempted re-fetch from the live \texttt{wikiart.org} site by title-matching succeeded for only $235/320$ ($73\%$), mostly because of title-encoding mismatches and distinct works sharing a title. The recorded filenames avoid this ambiguity.

\subsection{Source limitations}
Several properties of the source shape both the data and what the movement categories represent. WikiArt's labels are crowd-sourced rather than expert-assigned, and it is imbalanced across artists and movements. Our criterion of strongest association is not neutral, since it reflects which artists were historically collected, exhibited and digitised. Our own choices participate in it. Excluding Kahlo (Section~\ref{sec:data-margins}), for instance, removes one of the few non-Western-centred artists associated with Surrealism, on grounds of contested membership. The selection therefore captures the conventional canon of each movement rather than an objective or complete membership.

\subsection{Margin decisions}
\label{sec:data-margins}
Four selection decisions warrant note. \emph{(i)~Kahlo excluded from Surrealism}, because although Breton championed her, she rejected the label~\cite{babbs2024kahlo} and is widely read as closer to folk or magic realism, so we selected Tanguy, Delvaux, Masson and Carrington to keep the class to undisputed members. \emph{(ii)}~Three canonical Abstract Expressionists replaced, since de Kooning, Motherwell and Gottlieb are absent from this release. We added Newman, Still and Hofmann instead, all first-generation New York School figures. \emph{(iii)~Abstract Expressionism merges three sub-styles}, selected by artist, so no second-generation post-painterly painters enter. \emph{(iv)~Cubism merges three sub-styles}, with Robert Delaunay dropped (his work is tagged Orphism, a distinct movement not covered by the merge) and Louis Marcoussis replacing him.


\section{Method}
\label{sec:method}

\paragraph{Embeddings.}
Every painting is encoded once by a frozen, pretrained image encoder and reused across all analyses, with no fine-tuning. Each image is opened in RGB, passed through the encoder's standard preprocessing, and embedded with gradients disabled. The feature vector is $L_2$-normalised, so cosine similarity reduces to an inner product. Unless stated otherwise, results use CLIP ViT-B/32~\cite{radford2021clip} as the reference encoder, and Section~\ref{sec:backbones} repeats the full pipeline with CLIP ViT-L/14, the self-supervised DINOv2 ViT-B/14~\cite{oquab2023dinov2} (CLS token), and a supervised ImageNet ResNet-50~\cite{he2016resnet} (global-average-pooled). All nearest-neighbour computations use cosine distance.

\paragraph{Full-pool style classification.}
As a descriptive measure of how the dataset clusters by style, we run 5-NN classification over all 320 works, assigning each work the majority style among its five nearest neighbours excluding itself, with ties broken by the closest neighbour then alphabetically. We use $k = 5$ throughout, and accuracy is stable across $k \in \{3,5,7,9\}$. As a classifier-free view we report mean silhouette scores using the style labels as cluster assignments.

\paragraph{Artist-disjoint style classification.}
The full-pool measure, like standard random-split evaluation, permits other works by a query's own artist to serve as neighbours, so it cannot separate movement-level structure from individual-artist recognition. We control for this with a leave-one-artist-out protocol. Each of the 32 artists is held out in turn, that artist's ten works become queries, and the remaining 310 works form the reference pool, so no work is ever classified using another by the same painter. The protocol is exhaustive and deterministic. Each movement therefore yields eight accuracy figures, one for each held-out artist, and we report the standard deviation across them.

\paragraph{Artist-recognisability control.}
To test whether style success is partly artist recognition, we re-run the same procedure but predict the \emph{artist} rather than the style, with same-artist works permitted as neighbours, measuring how recognisable each artist is individually. We report overall accuracy against the $1/32$ chance rate, the mean per movement, and the per-artist values, set against the artist-disjoint style accuracies of the same artists.


\section{Results}

\subsection{Full-pool separability}
\label{sec:fullpool}

A 5-NN classifier over all 320 works recovers the four style labels with $0.869$ accuracy overall, well above the $0.25$ chance and majority-class rates. Per-style accuracy ranges from $0.975$ for Abstract Expressionism down to $0.713$ for Surrealism, with Cubism at $0.850$ and Impressionism at $0.938$. Silhouette scores are more cautious about cluster separation. The overall mean silhouette is only weakly positive ($0.066$), and Surrealism is the sole movement with a negative mean ($-0.039$), meaning many Surrealist works lie closer to other movements than to their own. These numbers are the baseline that the artist-disjoint protocol of Section~\ref{sec:loao} re-examines.

\FloatBarrier
\subsection{Artist-disjoint evaluation}
\label{sec:loao}

We now apply the artist-disjoint protocol of Section~\ref{sec:method}, holding each of the 32 artists out in turn.

\paragraph{Aggregate accuracy falls, unevenly across movements.}
Overall 5-NN style accuracy drops from $0.869$ to $0.766$ under the artist-disjoint protocol (Table~\ref{tab:loao}). That ten-point gap is the share of the full-pool score that depended on the model seeing other works by the query's own artist. The per-style breakdown shows the reliance is uneven. Impressionism and Cubism barely move, falling $3.8$ and $5.0$ points and remaining highly separable without same-artist neighbours. Surrealism falls $20.0$ points, from $0.713$ to $0.513$, the only movement for which the model misclassifies close to half of all held-out works. Abstract Expressionism sits between, dropping $12.5$ points while staying high at $0.850$.

\begin{table}[t]
\centering
\caption{Style classification accuracy under the full-pool protocol (all works eligible as neighbours) and the artist-disjoint protocol (each artist held out in turn). The last column is the standard deviation across the eight held-out artists of each movement.}
\label{tab:loao}
\setlength{\tabcolsep}{4pt}
\renewcommand{\arraystretch}{0.85}
\begin{tabular}{@{}lcccc@{}}
\toprule
Style & Full-pool & Artist-disjoint & Drop & std (across artists) \\
\midrule
Abstract Expressionism & 0.975 & 0.850 & 0.125 & 0.250 \\
Cubism & 0.850 & 0.800 & 0.050 & 0.187 \\
Impressionism & 0.938 & 0.900 & 0.038 & 0.132 \\
Surrealism & 0.713 & 0.513 & 0.200 & 0.226 \\
\midrule
Overall & 0.869 & 0.766 & 0.103 & --- \\
\bottomrule
\end{tabular}
\end{table}

\begin{figure}[t]
  \centering
  \includegraphics[width=0.7\textwidth]{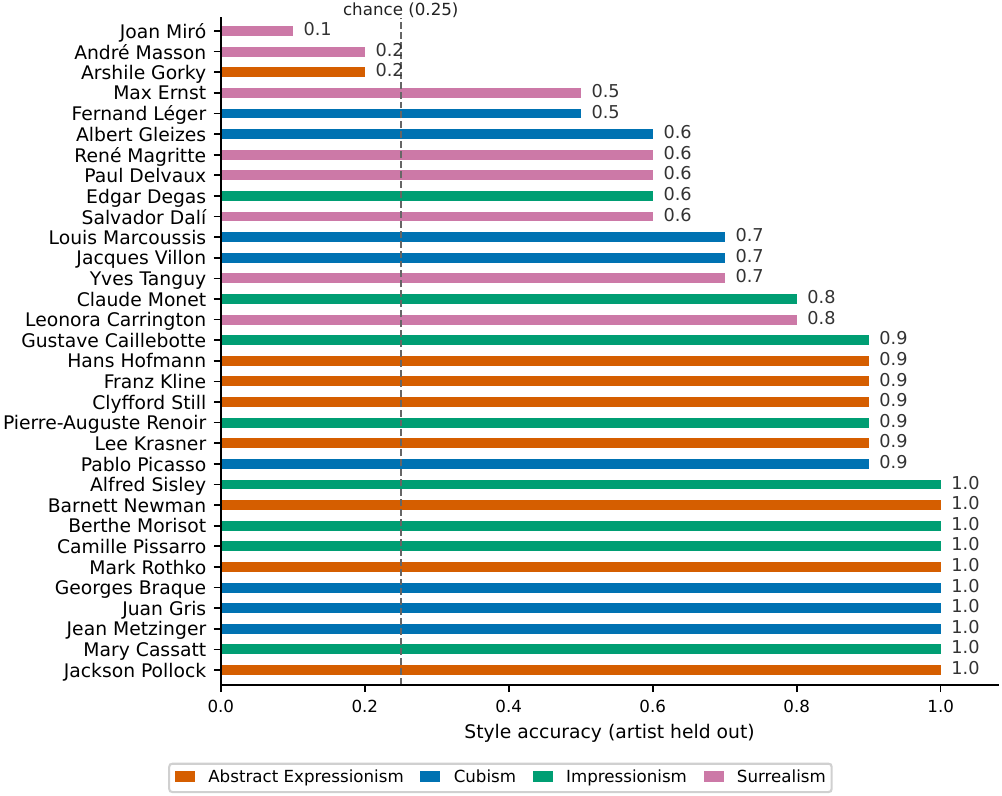}
  \caption{Per-artist style accuracy under the artist-disjoint protocol, all 32 artists sorted ascending and coloured by movement. The dashed line marks the four-class chance rate (0.25). The lowest scores are the automatist Surrealists Mir\'o (0.1) and Masson (0.2), with Gorky (0.2, Abstract Expressionism) the only non-Surrealist among them.}
  \label{fig:per-artist}
\end{figure}

\paragraph{Stable movements cohere through shared form, unstable ones through individual painters.} With only eight artists per movement, a single held-out painter can move a per-style figure substantially, so we report the standard deviation across the eight artists (Table~\ref{tab:loao}, last column). Impressionism has both the smallest drop and the smallest variance ($0.132$), the signature of a genuinely shared style. Surrealism combines the largest drop with high variance ($0.226$), and the per-artist view (Figure~\ref{fig:per-artist}) shows why. Its collapse is concentrated in particular artists rather than spread evenly. Two automatist painters, Mir\'o and Masson, fall to $0.1$ and $0.2$ when their own works are removed, while the other six Surrealists stay between $0.5$ and $0.8$. Surrealism's fragility is thus driven by a minority of its artists, which Section~\ref{sec:mechanism} takes up.

\paragraph{Errors concentrate in two destinations.} Abstract Expressionism and Cubism each receive $34$ misclassified works while emitting $12$ and $16$. Surrealism's $39$ errors go $18$ to Abstract Expressionism, $18$ to Cubism and $3$ to Impressionism.

\subsection{The mechanism: artist recognisability}
\label{sec:mechanism}

Section~\ref{sec:loao} showed that holding artists out lowers style accuracy unevenly and that Surrealism's fall is concentrated in particular painters. To ask what those painters have in common, we run the same nearest-neighbour procedure but predict the \emph{artist} rather than the style, with same-artist works permitted as neighbours. This measures how recognisable each artist is as an individual. If the embedding encodes who painted a work, a classifier can reach the correct movement by way of the artist rather than the style.

\paragraph{The artist signal is strong.} Across the 320 works, 5-NN artist classification reaches $0.466$ against a chance rate of $1/32 = 0.031$, identifying the individual painter roughly fifteen times more often than chance. Artist identity is therefore heavily represented and available to inflate style scores wherever same-artist neighbours are present.

\paragraph{Recognisability and style robustness are not simply related.} One might expect a clean inverse relationship, the more recognisable an artist the more their style score should collapse when held out. Across the 32 artists this relationship is essentially absent (Pearson $r = 0.06$). All four combinations occur. Some artists are recognisable and classify well by style, some are recognisable and collapse, some are unrecognisable yet classify well, and some are weak on both.

\paragraph{Three types of artist.}
The artists who drive Surrealism's collapse do not form one category. \emph{Memorised} artists are recognisable yet cannot be placed by style once held out. Gorky is the clearest case, recognised as himself at $0.90$ but classified by style at $0.20$. Mir\'o is a weaker case, recognised at $0.50$ but classified by style at $0.10$. For both, the embedding encodes the painter, and their earlier style accuracy was likely reaching the movement through the artist rather than through shared form. \emph{Poorly encoded} artists are weak on both measures. Masson is recognised at only $0.30$ and classified by style at $0.20$, so his low accuracy, unlike Mir\'o's, is not a memorised signal being taken away but a basic difficulty the embedding has with his work. The third type, the artist subsumed by a shared form, is best seen at the level of movements.

\paragraph{Shared-form movements subsume the individual.}
Aggregating recognisability by movement (Table~\ref{tab:recog}) produces a contrast that Section~\ref{sec:backbones} shows holds across encoders. Cubism has the lowest mean artist-recognisability at $0.24$. This is unsurprising, since Picasso and Braque's canvases from the Analytical phase are famously difficult to attribute~\cite{britannica_analyticalcubism}. Yet Cubism is among the most robust movements under hold-out, and Impressionism behaves similarly. Surrealism is the mirror image, its artists more individually recognisable ($0.51$) yet its style the least robust.

\begin{table}[t]
\centering
\caption{Mean artist-recognisability per movement (5-NN artist classification, same-artist neighbours permitted) against artist-disjoint style accuracy. Cubism and Impressionism combine low recognisability with high robustness; Surrealism combines higher recognisability with the lowest robustness.}
\label{tab:recog}
\setlength{\tabcolsep}{4pt}
\renewcommand{\arraystretch}{0.85}
\begin{tabular}{@{}lcc@{}}
\toprule
Movement & Artist-recognisability & Artist-disjoint style acc. \\
\midrule
Cubism & 0.238 & 0.800 \\
Impressionism & 0.400 & 0.900 \\
Abstract Expressionism & 0.713 & 0.850 \\
Surrealism & 0.513 & 0.513 \\
\bottomrule
\end{tabular}
\end{table}

\paragraph{A note on Abstract Expressionism.}
Abstract Expressionism fits neither of these patterns cleanly. Its artists are the most individually recognisable of any movement ($0.71$), yet the movement stays fairly robust under hold-out ($0.85$). Distinctive individual styles and a coherent movement-level signal can coexist, since Pollock, Rothko and Newman are each immediately recognisable and each classify perfectly under hold-out. The exception is Gorky, whose low style accuracy is hidden at the movement level by his robust colleagues.

\subsection{Generalisation across encoders}
\label{sec:backbones}

To test whether the pattern reflects an idiosyncrasy of CLIP ViT-B/32 or of image--text training, we repeat the full pipeline across four encoders differing in scale, architecture, and objective, the two CLIP models ViT-B/32 and ViT-L/14, the self-supervised vision-only DINOv2 ViT-B/14, and a supervised ImageNet ResNet-50. We compare patterns rather than absolute accuracies, since the encoders differ in training data, objective and architecture.

\paragraph{The fragility is encoder-general.} Surrealism is the most fragile movement under the artist-disjoint protocol for every encoder (Table~\ref{tab:backbones}), $0.513$, $0.638$, $0.375$, and $0.500$. The effect survives a change of scale within the CLIP family, where the larger ViT-L/14 lifts every accuracy but leaves Surrealism lowest, a change of architecture from transformer to convolutional network, and the move to a supervised encoder that is neither contrastive nor self-supervised. No encoder we test finds in Surrealism the shared visual form present in the other movements.

\begin{table}[t]
\centering
\caption{Per-style accuracy under the artist-disjoint protocol across four encoders. Absolute accuracies are not comparable across encoders; the consistent pattern is that Surrealism is lowest in every row (\textbf{bold}), while Impressionism and Cubism remain robust.}
\label{tab:backbones}
\setlength{\tabcolsep}{4pt}
\renewcommand{\arraystretch}{0.85}
\begin{tabular}{@{}lcccc@{}}
\toprule
Encoder & Abs.\ Expr. & Cubism & Impressionism & Surrealism \\
\midrule
CLIP ViT-B/32 & 0.850 & 0.800 & 0.900 & \textbf{0.513} \\
CLIP ViT-L/14 & 0.950 & 0.838 & 0.925 & \textbf{0.638} \\
DINOv2 ViT-B/14 & 0.913 & 0.763 & 0.875 & \textbf{0.375} \\
ResNet-50 (ImageNet) & 0.538 & 0.763 & 0.813 & \textbf{0.500} \\
\bottomrule
\end{tabular}
\end{table}

\paragraph{The effect is visual, not linguistic.}
DINOv2 is trained by self-supervision on images alone, with no text encoder, so it cannot exploit any correspondence between a painting and a verbal description of its style. If the Surrealism collapse arose from CLIP's language side, DINOv2 should not reproduce it. It reproduces it more sharply than any CLIP model, with Surrealism at $0.375$. The fragility is therefore in the visual structure of the movements, not in language-derived associations. The movement-level relationship of Section~\ref{sec:mechanism} also recurs in every encoder. Cubism and Impressionism pair the lowest artist-recognisability with robust hold-out accuracy, and Surrealism pairs higher recognisability with the lowest hold-out accuracy. Under DINOv2 the order reverses on recognisability, with Impressionism marginally below Cubism.


\section{Discussion}
\label{sec:discussion}

\paragraph{Movements are separable for different reasons.} Our results argue against reading a single style-classification accuracy as a measure of how well an embedding ``understands'' style. The aggregate figure of $0.87$ conceals a range from $0.51$ to $0.90$ once artists are held out. Impressionism and Cubism are held together by visual form shared across their members and survive the removal of any single artist, whereas Surrealism coheres only loosely, with much of its apparent separability resting on the embedding recognising particular painters. ``Does the embedding capture style?'' is therefore the wrong question, since it presupposes that style works the same way in every movement. The better question is which movements remain separable once individual authorship is controlled for, and on what basis.

\paragraph{Shared form versus artist lookup.}
The artist-recognisability control names the two ends of this spectrum. At one end, a movement classifies robustly because its artists are hard to tell apart. Cubism has the lowest artist-recognisability of any movement yet among the highest robustness under hold-out. The encoder cannot reliably identify the painter yet still places the work correctly, which rebuts the suspicion that a style classifier is merely an artist classifier in disguise. At the other end sits Surrealism, whose separability collapses once particular painters are removed (Section~\ref{sec:mechanism}). We do not read this as a defect of the encoder. Surrealist practice prized idiosyncratic personal languages, so an artist's individual brand of Surrealism may be the most stylistic thing there is to learn, and in such a movement detecting the artist and detecting the style partly collapse into one task. That Impressionists resemble one another more than Surrealists do is not a flaw in the comparison but the difference the protocol exists to measure. The problem lies in the evaluation, since a random split reports the resulting accuracy as movement-level competence, transferable to unseen artists, which for Surrealism it is not. The artist-disjoint protocol does not create that collapse but reveals it, and quantifies how far each movement's label is recoverable from shared form alone.

\paragraph{Movements differ in what unifies them.}
That the same ordering appears in every encoder we test, including a vision-only one with no access to text (Section~\ref{sec:backbones}), indicates a property of the movements themselves rather than an artefact of one model or of language. This fits art-historical intuition. Impressionism and Cubism are defined substantially by how their surfaces look, broken colour and dissolved light in one case and the fragmentation of form into geometric facets in the other, whereas Surrealism is unified far more by a shared idea, the depiction of dream logic and incongruous juxtaposition, than by any shared appearance.

\paragraph{Implications for evaluation.}
Random-split evaluation can substantially overstate how well a model captures style, by up to twenty points for one movement here. The overstatement is uneven across movements, so the aggregate hides where it falls. Because these embeddings are widely used off the shelf, an evaluation that allows same-artist works on both sides of the split risks certifying as ``style understanding'' what is partly artist recognition. Artist-disjoint evaluation is inexpensive, needs no retraining, and turns a brittle aggregate into an interpretable per-movement and per-artist picture, so we suggest it as a default rather than a refinement when the question is genuinely about style. The principle is not novel, since group-disjoint evaluation is standard in domains such as medical imaging and speaker-independent speech recognition. Our contribution is to bring it to painting-style classification, where a recent survey observes it has been largely absent~\cite{styleclass_survey2026}, and to show how much it changes the conclusions.

\paragraph{Two cautions.} First, we lean only on the ordering of movements within each encoder, since absolute accuracies are not comparable across them. One departure is that the ImageNet-trained ResNet-50 classifies Abstract Expressionism far less robustly than the transformer encoders, plausibly because a convolutional network trained for object recognition relies on local texture, and gestural abstraction offers texture without the objects such a model expects. Second, our dataset is small and its movements are not matched for subject matter, so some of what we attribute to style remains entangled with content and composition, a limitation taken up in Section~\ref{sec:limitations}. The central result is the fall from full-pool to artist-disjoint accuracy rather than either figure alone, which neither caution touches.


\section{Limitations}
\label{sec:limitations}

\paragraph{Scale and balance.}
The style dataset is small, 320 works across 32 artists. A single held-out painter moves a per-style figure by as much as $0.125$, which is why we report the spread across held-out artists alongside the mean, and point to individual artists rather than to confidence intervals. Having equal numbers of artists and works per movement aids interpretation, but the absolute accuracies are specific to this curated set and should not be read as what these encoders would achieve on paintings generally.

\paragraph{Content is not controlled.} The movements are not matched for subject matter or composition, so style remains entangled with content. Impressionist landscapes and Cubist still lifes, for example, differ in what they depict as well as how. Some of what we attribute to style separability may therefore reflect subject and composition. This does not undermine the comparative result, since the confound applies across movements and cannot by itself explain why holding out artists costs Surrealism twenty points and Impressionism four, but it does mean the per-movement accuracies should not be read as measuring style in isolation.

\paragraph{Labels, canon, and preprocessing.} The dataset inherits its source's properties, crowd-sourced rather than expert-assigned labels and a selection reflecting the conventional Western and male-skewed canon (Section~\ref{sec:data}). Our preprocessing adds anisotropic resizing, which distorts aspect ratio uniformly across movements and encoders, though we cannot rule out that it affects geometry-dependent movements more than others.

\paragraph{Scope of the probe.}
We study frozen embeddings used as fixed feature extractors with nearest-neighbour classification, a deliberately simple and transparent setting. We do not fine-tune, probe intermediate layers, or use the encoders' text towers. Our claims therefore concern what off-the-shelf image embeddings reveal under nearest-neighbour comparison, not the limits of what these models could represent after adaptation. Fine-tuning on art-specific labels is known to improve stylistic sensitivity~\cite{ghildyal2025wpclip}, and how artist dependence behaves under such adaptation is left to future work.

\section{Conclusion}
\label{sec:conclusion}

Frozen vision embeddings classify paintings by style with high accuracy, but that result has typically been measured under evaluations that let works by the same artist appear on both sides of the split. Holding out whole artists lowers 5-NN style accuracy from $0.87$ to $0.77$, and the modest aggregate change conceals a sharply uneven effect: Impressionism and Cubism are essentially unaffected, while Surrealism falls twenty points. An artist-recognisability control accounts for the unevenness. The robust movements are those whose artists the embedding is least able to distinguish, indicating a genuine shared visual form that has subsumed individual authorship, whereas Surrealism's separability rests substantially on the recognisability of particular painters and collapses when they are removed. The pattern holds across four encoders differing in scale, architecture, and objective, including a vision-only model whose agreement places the effect in visual structure rather than language.

The broader point is methodological. A split that lets a classifier reach the movement through the artist tests something other than style, and holding whole artists out is enough to block that route. Natural extensions include larger and less curated datasets spanning more movements and artists, classifiers beyond nearest neighbour, content-controlled subsets that match subject across movements, probing the text encoders alongside the image ones, and asking whether fine-tuning closes or merely relocates the artist dependence we have measured.

\bibliographystyle{splncs04}
\bibliography{main}
\end{document}